\documentclass[sn-basic,square,comma,numbers,iicol]{sn-jnl}
\setcitestyle{numbers}

\makeatletter
\renewcommand\@biblabel[1]{#1.}
\makeatother
\usepackage{array}
\newcolumntype{P}[1]{>{\centering\arraybackslash}p{#1}}

\usepackage[retain-explicit-plus]{siunitx}
\usepackage{makecell}
\usepackage{lineno}
\usepackage{hyperref}
\usepackage{multirow}
\usepackage{xcolor}
\usepackage{mathtools}
\usepackage{subcaption}
\usepackage{comment}
\usepackage{graphicx}
\usepackage{amsmath,amssymb,amsfonts}
\usepackage{amsthm}
\usepackage{mathrsfs}
\usepackage[title]{appendix}
\usepackage{textcomp}
\usepackage{manyfoot}
\usepackage{booktabs}
\usepackage{algorithm}
\usepackage{algorithmicx}
\usepackage{algpseudocode}
\usepackage{listings}
\usepackage{pifont}
\newcommand{\cmark}{\ding{51}}
\newcommand{\xmark}{\ding{55}}
\usepackage{lineno}
\DeclarePairedDelimiter{\nint}\lfloor\rceil

\usepackage{blindtext}
\usepackage{hyperref}
\usepackage{nameref}

\newcounter{mylabelcounter}

\makeatletter
\newcommand{\labelText}[2]{
\refstepcounter{mylabelcounter}
\immediate\write\@auxout{
 \string\newlabel{#2}{{\unexpanded{#1}}{\thepage}{{\unexpanded{#1}}}{mylabelcounter.\number\value{mylabelcounter}}{}}
}
}
\makeatother

\begin{document}

\title[]{Leveraging Large Language Models through Natural Language Processing to provide interpretable Machine Learning predictions of mental deterioration in real time}

\author[1]{\fnm{Francisco} \sur{de Arriba-Pérez}}\email{farriba@gti.uvigo.es}
\equalcont{These authors contributed equally to this work.}

\author*[1]{\fnm{Silvia} \sur{García-Méndez}}\email{sgarcia@gti.uvigo.es}
\equalcont{These authors contributed equally to this work.}

\affil[1]{\orgname{Information Technologies Group, atlanTTic, University of Vigo}, \city{Vigo}, \country{Spain}}

\abstract{Based on official estimates, 50 million people worldwide are affected by dementia, and this number increases by 10 million new patients every year. Without a cure, clinical prognostication and early intervention represent the most effective ways to delay its progression. To this end, Artificial Intelligence and computational linguistics can be exploited for natural language analysis, personalized assessment, monitoring, and treatment. However, traditional approaches need more semantic knowledge management and explicability capabilities. Moreover, using Large Language Models (\textsc{llm}s) for cognitive decline diagnosis is still scarce, even though these models represent the most advanced way for clinical-patient communication using intelligent systems. Consequently, we leverage an \textsc{llm} using the latest Natural Language Processing (\textsc{nlp}) techniques in a chatbot solution to provide interpretable Machine Learning prediction of cognitive decline in real-time. Linguistic-conceptual features are exploited for appropriate natural language analysis. Through explainability, we aim to fight potential biases of the models and improve their potential to help clinical workers in their diagnosis decisions. More in detail, the proposed pipeline is composed of (\textit{i}) data extraction employing \textsc{nlp}-based prompt engineering; (\textit{ii}) stream-based data processing including feature engineering, analysis, and selection; (\textit{iii}) real-time classification; and (\textit{iv}) the explainability dashboard to provide visual and natural language descriptions of the prediction outcome. Classification results exceed \SI{80}{\percent} in all evaluation metrics, with a recall value for the mental deterioration class about \SI{85}{\percent}. To sum up, we contribute with an affordable, flexible, non-invasive, personalized diagnostic system to this work.}

\keywords{Artificial Intelligence, explainability, healthcare, Large Language Models, Natural Language Processing, stream-based Machine Learning.}

\maketitle

\textbf{This version of the article has been accepted for publication, after peer review but is not the Version of Record and does not reflect post-acceptance improvements, or any corrections. The Version of Record is available online at: https://doi.org/10.1007/s13369-024-09508-2.}

\section{Introduction}
\label{sec:introduction}

Neurodegenerative Alzheimer’s disorder (\textsc{ad}) is the leading cause of chronic or progressive dementia, which negatively impacts cognitive functioning, including comprehension, speech, and thinking problems, memory loss, etc. \citep{Vats2021}. More in detail, the typical stages of cognitive decline can be categorized as pre-clinical \textsc{ad}, Mild Cognitive Impairment (\textsc{mci}) caused by \textsc{ad}, and finally \textsc{ad} dementia \citep{mao2023ad}. Generally, cognitively impaired users find difficult to perform daily tasks with the consequent detrimental impact on their life quality \citep{Syed2020}. In this line, cognitive decline is a leading cause of dependency and disability for our elders \citep{Nadira2020}.

According to the Alzheimer's Association report on the impact of this disease in the United States \citep{AlzheimersAssociation2023}, it is the sixth‐leading death cause that increased more than \SI{145}{\percent} in the last years. Moreover, it affects 6.7 million people 65 or older. Dreadfully, this number is predicted to grow to 13.8 million by 2060. Regarding medical expenses related to people affected with dementia 65 or older, these are three times greater than those of people without this condition, reaching 345 billion dollars so far in 2023. Overall, the World Health Organization estimates that 50 million people worldwide are affected by dementia, with 10 million new patients yearly\footnote{Available at \url{https://www.who.int/news-room/fact-sheets/detail/dementia}, May 2024.}. 

Clinical prognostication and early intervention, the most promising ways to address mental deterioration, rely on effective progression detection \citep{mao2023ad}. Among the benefits of early identification, care planning assistance, medical expense reduction, and the opportunity to receive the latest treatments, including non-invasive therapy, given the rapid biologic therapeutics advancements, stand out \citep{Rasmussen2019,Manly2021}. The social stigma and socioeconomic status must also be considered when accessing mental health services \citep{Kandratsenia2019}. However, the latter early diagnosis is challenging since the symptoms can be confused with normal aging decline \citep{Elliot1019}. To address it, computational linguistics can be exploited \citep{Pl2024}. Natural language analysis is particularly relevant, constituting a significant proportion of healthcare data \citep{Velupillai2018}. 
Particularly, impairment in language production mainly affects lexical (\textit{e.g.}, little use of nouns and verbs), semantic (\textit{e.g.}, the use of empty words like thing/stuff), and pragmatic (\textit{e.g.}, discourse disorganization) aspects \citep{Yuan2020}.

Digital and technological advances such as Artificial Intelligence (\textsc{ai})-based systems represent promising approaches towards individuals' needs for personalized assessment, monitoring, and treatment \citep{bertacchini2023social}. Accordingly, these systems have the capabilities to complement traditional methodologies such as the Alzheimer’s Disease Assessment Scale-Cognition (\textsc{adasc}og), the Mini-Mental State Examination (\textsc{mmse}), and the Montreal Cognitive Assessment (\textsc{moca}), which generally involve expensive, invasive equipment, and lengthy evaluations \citep{agbavor2022predicting}. In fact, paper-and-pencil cognitive tests continue to be the most common approaches even though the latest advances in the Natural Language Processing (\textsc{nlp}) field enable easy screening from speech data while at the same time avoiding patient/physician burdening \citep{Chen2021}. Summing up, language analysis can translate into an effective, inexpensive, non-invasive, and simpler way of monitoring cognitive decline \citep{agbavor2022predicting,Li2022} provided that spontaneous speech of cognitive impaired people is characterized by the aforementioned semantic comprehension problems and memory loss episodes \citep{Santander2022}.

Consequently, Clinical Decision Support Systems (\textsc{cdss}s), Diagnostic Decision Support Systems (\textsc{ddss}s), and Intelligent diagnosis systems (\textsc{ids}s) which apply \textsc{ai} techniques (\textit{e.g.}, Machine Learning - \textsc{ml}, \textsc{nlp}, etc.) to analyze patient medical data (\textit{i.e.}, clinical records, imaging data, lab results, etc.) and discover relevant patterns effectively and efficiently, have significantly attracted the attention of the medical and research community \citep{Caruccio2023}. However, one of the main disadvantages of traditional approaches is their lack of semantic knowledge management and explicability capabilities \citep{Santander2022}. The latter can be especially problematic in the medical domain regarding accountability of the decision process for the physicians to recommend personalized treatments \citep{agbavor2022predicting}. 

Integrating \textsc{ai}-based systems in conversational assistants to provide economical, flexible, immediate, and personalized health support is particularly relevant \citep{Prasad2023}. Their use has been greatly enhanced by the nowadays popular Large Language Models (\textsc{llm}s), enabling dynamic dialogues compared to previous developments \citep{Palanica2019}. Subsequently, \textsc{llm}s have been powered by the latest advancements in deep learning techniques and the availability of vast amounts of cross-disciplinary data \citep{Idris2024}. These models represent the most innovative approach of \textsc{ai} into healthcare by expediting medical interventions and providing new markers and therapeutic approaches to neurological diagnosis from patient narrative processing \citep{romano2023large}. Note that patient experience can also be improved with the help of \textsc{llm}s in terms of information and support seeking \citep{Fear2023}. Summing up, conversation assistants that leverage \textsc{llm}s have the potential to monitor high-risk populations and provide personalized advice, apart from offering companion \citep{Prasad2023,Alessa2023} constituting the future of therapy in the literature \citep{Vaidyam2019}.

Given the still poor accuracy of \textsc{cdss}s \citep{Ceney2021,Schmieding2022}, we plan to leverage an \textsc{llm} using the latest \textsc{nlp} techniques in a chatbot solution to provide interpretable \textsc{ml} prediction of cognitive decline in real-time. Linguistic-conceptual features are exploited for appropriate natural language analysis. The main limitation of \textsc{llm}s is that their outcomes may be misleading. Thus, we apply prompt engineering to avoid the ``hallucination" effect. Through explainability, we aim to fight potential biases of the models and improve their potential to help clinical workers in their diagnosis decisions. Summing up, we contribute with an affordable, non-invasive diagnostic system in this work. 

The rest of this paper is organized as follows. Section \ref{sec:related_work} reviews the relevant competing works on cognitive decline detection involving \textsc{llm}s and interpretable \textsc{ml} predictions of mental deterioration. The contribution of this work is summarized in Section \ref{sec:contributions}. Section \ref{sec:methodology} explains the proposed solution, while Section \ref{sec:results} describes the experimental data set, our implementations, and the results obtained. Finally, Section \ref{sec:conclusions} concludes the paper and proposes future research.

\begin{description}
    \item \textbf{Problem}. The World Health Organization predicts a yearly increase of 10 million people affected with dementia.
    \item \textbf{What is already known}. Paper-and-pencil cognitive tests continue to be the most common approach. The latter is impractical, given the disease growth rate. Moreover, one of the main disadvantages of intelligent approaches is their lack of semantic knowledge management and explicability capabilities.
    \item \textbf{What this paper adds}. We leverage an \textsc{llm} using the latest \textsc{nlp} techniques in a chatbot solution to provide interpretable \textsc{ml} prediction of cognitive decline in real-time. To sum up, we contribute with an affordable, flexible, non-invasive, personalized diagnostic system to this work.
\end{description}

\section{Related work}
\label{sec:related_work}

As previously mentioned, the main focus of dementia treatment is to delay the cognitive deterioration of patients \citep{Santander2022}. Consequently, early diagnosis, which simultaneously contributes to reducing medical expenses in our aging society and avoiding invasive treatments with subsequent side effects on the users, is desirable \citep{Rasmussen2019}. To this end, \textsc{ai} has been successfully applied to \textsc{ids}s in order to recommend treatments based on their diagnosis prediction \citep{Serhat2023,Yu2023}. 

While \textsc{ml} models perform well and fast in diagnosis tasks, they require extensive training data previously analyzed by experts, which is labor-intensive and time-consuming \citep{Santander2022}. In contrast, advanced \textsc{nlp}-based solutions exploit transformer-based models already trained with large corpora, including domain-related data, which results in very sensitive text analysis capabilities \citep{Koga2023}. Consequently, transformer-based pre-trained language models (\textsc{plm}s) (\textit{e.g.}, \textsc{bert} \citep{kenton2019bert}, \textsc{gpt-3} \citep{gpt3-2020}) which preceded the popular \textsc{llm}s (\textit{e.g.}, \textsc{gpt-4}\footnote{Available at \url{https://platform.openai.com/docs/models/gpt-4}, May 2024.}) have disruptively transformed the \textsc{nlp} research. These models exhibit great contextual latent feature extraction abilities from textual input \citep{Koga2023}. The latter models are implemented to predict the next token based on massive training data, resulting in a word-by-word outcome \citep{Deriu2021}. Nowadays, they are used for various tasks, including problem-solving, question-answering, sentiment analysis, text classification, and generation, etc. \citep{brown2020language}. 

There exist \textsc{plm} versions over biomedical and clinical data such as Bio\textsc{bert} \citep{Lee2020}, Bio\textsc{gpt} \citep{Luo2022}, Blue\textsc{bert} \citep{Peng2019}, Clinical\textsc{bert}\footnote{Available at \url{https://github.com/EmilyAlsentzer/clinicalBERT}, May 2024.} and \textsc{tcm-bert} \citep{Yao2019}. Open-domain conversational assistants, whose dialogue capabilities are not restricted to the conversation topic, exploit \textsc{llm}s \citep{Prasad2023}. However, using \textsc{llm}s for cognitive decline diagnosis is still scarce even though these models represent the most advanced way for clinical-patient communication using intelligent systems \citep{Hirosawa2023}. More in detail, they overcome the limitation of traditional approaches that lack semantic reasoning, especially relevant in clinical language \citep{Gillioz2020}. Unfortunately, despite the significant advancement they represent, \textsc{llm}s still exhibit certain limitations in open-domain task-oriented dialogues (\textit{e.g.}, medical use cases) \citep{Ji2023}. For the latter, the Reinforcement Learning from Human Feedback (\textsc{rlhf}, \textit{i.e.}, prompt engineering) technique is applied to enhance their performance based on end users' instructions and preferences \citep{Chen2023}. 

Regarding the application of \textsc{plm} to the medical field, \citet{Syed2020} performed two tasks: (\textit{i}) dementia prediction and (\textit{ii}) \textsc{mmse} score estimation from speech recordings combining acoustic features and text embeddings obtained with the \textsc{bert} model from their transcription. The input data correspond to cognitive tests (\textsc{ct}s). \citet{Yuan2020} analyzed disfluencies (\textit{i.e.}, uh/um word frequency and speech pauses) with \textsc{bert} and \textsc{ernie} modes based on data from the Cookie Theft picture from the Boston Diagnostic Aphasia Exam. Close to the work by \citet{Syed2020}, \citet{Chen2021} analyzed the performance of \textsc{bert} model to extract embeddings in cognitive impairment detection from speech gathered during \textsc{ct}s. \citet{Santander2022} combined the Siamese \textsc{bert} networks (\textsc{sbert}s) with \textsc{ml} classifiers to firstly extract the sentence embeddings and then predict Alzheimer’s disease from \textsc{ct} data. In contrast, \citet{Vats2021} performed dementia detection combining \textsc{ml}, the \textsc{bert} model, and acoustic features to achieve improved performance. Moreover, \citet{Li2022} compared \textsc{gpt-2} with its artificially degraded version (\textsc{gpt-d}) created with a dementia-related linguistic anomalies layer induction based on data from a picture description task, while \citet{agbavor2022predicting} predicted dementia and cognitive score from \textsc{ct} data using \textsc{gpt-3} exploiting both word embeddings and acoustic knowledge. Finally, \citet{mao2023ad} pre-trained the \textsc{bert} model with unstructured clinical notes from Electronic Health Records (\textsc{ehr}s) to detect \textsc{mci} to \textsc{ad} progression.

More closely related to our research is the work by \citet{bertacchini2023social}. The authors designed Pepper, a social robot with real-time conversational capabilities exploiting the Chat\textsc{gpt} \textsc{gpt-3.5} model. However, the use case of the system is Autism Spectrum Disorder detection. Furthermore, \citet{Caruccio2023} compared the diagnoses performance of different models of Chat\textsc{gpt} (\textit{i.e.}, \textsc{ada}, \textsc{babbage}, \textsc{curie}, \textsc{davinci} and \textsc{gpt-3.5}) with Google Bard and traditional \textsc{ml} approaches based on symptomatic data. The authors exploited prompt engineering to ensure appropriate performance when submitting clinical-related questions to the \textsc{llm} model. Moreover, \citet{Hirosawa2023} analyzed the diagnosis ability of Chat\textsc{gpt} \textsc{gpt-3.5} model using clinical vignettes. Then, the \textsc{llm} was evaluated compared to physicians’ diagnosis. However, the authors again focus not on cognitive decline prediction but on ten common chief complaints. Consideration should be given to the work by \citet{Koga2023}, who used Chat\textsc{gpt} (\textit{i.e.}, \textsc{gpt-3.5} and \textsc{gpt-4} models) and Google Bard to predict several neurodegenerative disorders based on clinical summaries in clinicopathological conferences without being a specific solution tailored for \textsc{ad} prediction. Finally, regarding conversational assistants that integrate \textsc{llm}s, \citet{Zaman2023} developed a chatbot based on Chat\textsc{gpt} \textsc{gpt-3.5} model to provide emotional support to caregivers (\textit{i.e.}, practical tips and shared experiences).

\subsection{Contributions}
\label{sec:contributions}

As previously described, a vast amount of work in the state of the art exploits \textsc{plm}s even in the clinical field \citep{Alomari2022}. However, scant research has been performed in the case of \textsc{llm} models. Table \ref{tab:comparison} summarizes the reviewed diagnostic solutions that exploit \textsc{llm}s in the literature. Note that explainability represents a differential characteristic of the solution proposed given the relevance of promoting transparency in \textsc{ai}-based systems \citep{Wischmeyer2020}.

Given the comparison with competing works:
\begin{itemize}
 \item Our system is the first that jointly considers the application of an \textsc{llm} over spontaneous speech and provides interpretable \textsc{ml} results for the use case of mental decline prediction. 
 \item Our solution implements \textsc{ml} models in streaming to provide real-time functioning, hence avoiding the re-training cost of batch systems. 
 \item In this work, we leverage the potential of \textsc{llm}s by applying the \textsc{rlhf} technique through prompt engineering in a chatbot solution. Note that the natural language analysis is performed with linguistic-conceptual features. Consequently, we contribute with an affordable, non-invasive diagnostic system. 
 \item Our system democratizes access to researchers and end users within the public health field to the latest advances in \textsc{nlp}. 
\end{itemize}

\begin{table*}[!htbp]
\centering
\caption{Comparison of diagnostic \textsc{llm}-based solutions taking into account the field of application, the model used, the input data, and explainability (Ex.) capability.}
\label{tab:comparison}
\begin{tabular}{ccccc}
\toprule
\textbf{Authorship} & \textbf{Application} & \textbf{LLM} & \textbf{Input data} & \textbf{Ex.}\\
\midrule

\multirow{3}{*}{\citet{Caruccio2023}} & \multirow{3}{*}{General diagnosis} & Chat\textsc{gpt} & \multirow{3}{*}{Symptomatic data} & \multirow{3}{*}{\xmark} \\
& & Google Bard\\
& & \textsc{ml}\\

\citet{Hirosawa2023} & Common complaints & Chat\textsc{gpt} & Clinical vignettes & \xmark\\

\multirow{2}{*}{\citet{Koga2023}} & \multirow{2}{*}{Neurodegenerative disorders} & Chat\textsc{gpt} & \multirow{2}{*}{Clinical summaries} & \multirow{2}{*}{\xmark}\\
& & Google Bard\\
\midrule
\textbf{Proposal} & Mental decline & Chat\textsc{gpt} + \textsc{ml} & Spontaneous speech & \cmark\\
\bottomrule
\end{tabular}
\end{table*}

\section{Methodology}
\label{sec:methodology}

\begin{figure*}[!htpb]
 \centering
 \includegraphics[scale=0.129]{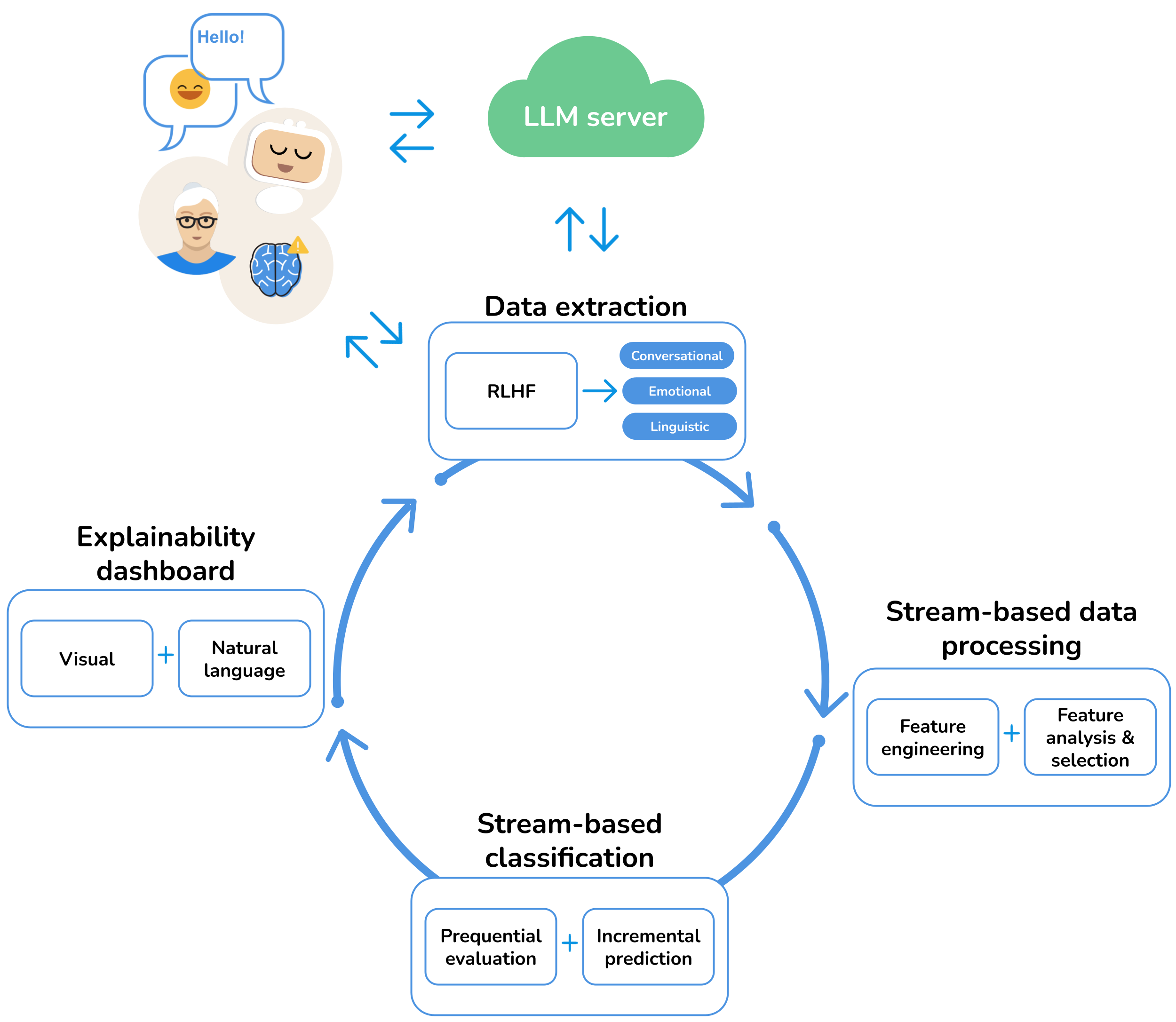}
 \caption{System scheme.}
 \label{fig:scheme}
\end{figure*}

Figure \ref{fig:scheme} depicts the system scheme proposed for real-time prediction of mental decline combining \textsc{llm}s and \textsc{ml} algorithms with explainability capabilities. More in detail, it is composed of (\textit{i}) data extraction employing \textsc{nlp}-based prompt engineering (Section \ref{sec:data_extraction}); (\textit{ii}) stream-based data processing including feature engineering, analysis and selection (Section \ref{sec:data_processing}); (\textit{iii}) real-time classification (Section \ref{sec:classification}); and (\textit{iv}) the explainability dashboard to provide visual and natural language descriptions of the prediction outcome (Section \ref{sec:explainability}). Algorithm \ref{alg:methodology} describes the complete process.

\begin{algorithm*}[!htbp]
 \caption{\label{alg:methodology}{\bf Methodology}}
 \begin{algorithmic}[0]
 
    \State $scenario, model\_name,selector\_mode,selector\_threshold$ \hfill{\%Configuration parameters defined by the user}

    \State $count=0$
    \State $list\_y, list\_y\_pred, list\_sessions=[]$
    \State $input(session)$ \hfill{\%A new dialogue session enters the system}
    \While{$session != null$}
        \State $list\_sessions.append(session)$
        \State $list\_features=data\_extraction(session)$
        \State $list\_features\_selected=data\_processing(list\_sessions,list\_features,selector\_mode,selector\_threshold)$
        \State $list\_y.append(session.y)$
        \State $y\_pred=classification(list\_features\_selected, scenario, model\_name, count, list\_y$)
        \State $list\_y\_pred.append(y\_pred)$
        \State $count=count+1$
        \State $input(session)$
     \EndWhile

 \end{algorithmic}
\end{algorithm*} 

\subsection{Data extraction}
\label{sec:data_extraction}

The Chat\textsc{gpt} \textsc{gpt-3.5} model used serves two purposes: (\textit{i}) it enables a natural, free dialogue with the end users, and (\textit{ii}) data is extracted due to its semantic knowledge management capabilities. The latter information is gathered once the conversation is concluded (either more than 3 minutes of inactivity or farewell detected) and used to compute the features used for classification (see Section \ref{sec:feature_engineering}). For this extraction, prompt engineering is exploited. The complete data extraction process is described in Algorithm \ref{alg:data_extraction}.

\begin{algorithm*}[!htbp]
 \caption{\label{alg:data_extraction}{\bf Data extraction}}
 \begin{algorithmic}[0] 
 \Function{$data\_extraction$}{$session$}

    \State $list\_human\_interactions=[]$ \hfill{\%To save only the human interactions, excluding those made by the chatbot}
    \State $complete\_human\_dialogue=``"$
    \For{$item$ $in$ $session$} 
        \If{$item.type()==``human"$}
            \State $list\_human\_intereactions.append(item)$
            \State $complete\_human\_dialogue=complete\_human\_dialogue.concat(item)$
        \EndIf
    \EndFor
    \State $feature\_9=len(list\_human\_intereactions)$ \hfill{\%See Table 2}
    \State $feature\_10=len(complete\_human\_dialogue.split())$
    \State $rest\_features=prompt\_data\_extraction()$ \hfill{\%See Listing 1}
    \State $return(feature\_9, feature\_10, rest\_features)$ 

 \EndFunction
 \end{algorithmic}
\end{algorithm*}

\subsection{Stream-based data processing}
\label{sec:data_processing}

Stream-based data processing encompasses feature engineering, analysis, and selection tasks to ensure the optimal performance of the \textsc{ml} classifiers.

\subsubsection{Feature engineering}
\label{sec:feature_engineering}

Table \ref{tab:features} details the features used to predict mental decline. Note that conversational, emotional, and linguistic-conceptual features are computed. The conversational features\footnote{Features 9-10 are not computed using the \textsc{llm}.} (1-10) represent relevant semantic and pragmatic information related to the free dialogue (\textit{e.g.}, fluency, repetitiveness, etc.), while emotional features focus on the mental and physical state of the users. Finally, linguistic features represent lexical and semantic knowledge (\textit{e.g.}, disfluencies, placeholder words, etc.).

Furthermore, the system maintains a history of each user data (\textit{i.e.}, past and current feature values) that enables the computation of four new characteristics per each in Table \ref{tab:features}: average, \textsc{q}1, \textsc{q}2, and \textsc{q}3 as indicated in Equation \eqref{eq:feature_engineering}, where $n$ is the user conversation counter and $X[n]$ represents a particular feature with historical data.

\begin{table*}[!htbp]
\centering
\footnotesize
\caption{\label{tab:features}Features engineered for mental deterioration prediction.}
\begin{tabular}{llll}
\toprule
\bf Category & \bf ID & \bf Name & \bf Description\\\hline

\multirow{10}{*}{Conversational} 

& 1 & Amnesia & \begin{tabular}[c]{@{}p{8cm}@{}} Showing difficulty in recalling past data.\end{tabular}\\

& 2 & Incoherence & \begin{tabular}[c]{@{}p{8cm}@{}} Use of inconsistent responses.\end{tabular}\\

& 3 & Incomprehension & \begin{tabular}[c]{@{}p{8cm}@{}} Inability to understand certain aspects.\end{tabular}\\

& 4 & Confusion & \begin{tabular}[c]{@{}p{8cm}@{}} Showing uncertainty about what is discussed.\end{tabular}\\

& 5 & Fluency & \begin{tabular}[c]{@{}p{8cm}@{}} Use of smooth quality utterances.\end{tabular}\\

& 6 & Initiative & \begin{tabular}[c]{@{}p{8cm}@{}} Willingness to engage in the dialogue even posing questions.\end{tabular}\\

& 7 & Repetitiveness & \begin{tabular}[c]{@{}p{8cm}@{}} Use of repetitive utterances that affect the conversation flow.\end{tabular}\\

& 8 & Secretive & \begin{tabular}[c]{@{}p{8cm}@{}} Inclined to hide feelings and personal information.\end{tabular}\\

& 9 & Interactions & \begin{tabular}[c]{@{}p{8cm}@{}} Total number of bot-human interaction pairs in the dialogue.\end{tabular}\\

& 10 & Words & \begin{tabular}[c]{@{}p{8cm}@{}} Total number of words in the dialogue.\end{tabular}\\

\midrule

\multirow{5}{*}{Emotional} 

& 11 & Health state & \begin{tabular}[c]{@{}p{8cm}@{}} Absence/presence of mental or physical health concerns.\end{tabular}\\

& 12 & Fatigue & \begin{tabular}[c]{@{}p{8cm}@{}} Sense of tiredness.\end{tabular}\\

& 13 & Loneliness & \begin{tabular}[c]{@{}p{8cm}@{}} Sense of abandonment.\end{tabular}\\

& 14 & Polarity & \begin{tabular}[c]{@{}p{8cm}@{}} Providing negative, neutral or positive information.\end{tabular}\\

& 15 & Sadness & \begin{tabular}[c]{@{}p{8cm}@{}} Sense of depression.\end{tabular}\\

\midrule

\multirow{7}{*}{Linguistic} 

& 16 & Colloquial registry & \begin{tabular}[c]{@{}p{8cm}@{}} Using a casual and simple language registry.\end{tabular}\\

& 17 & Conjugation problems & \begin{tabular}[c]{@{}p{8cm}@{}} Inability to correctly conjugate verb tenses.\end{tabular}\\

& 18 & Disfluency & \begin{tabular}[c]{@{}p{8cm}@{}} Use of interjections to complete pauses.\end{tabular}\\

& 19 & Formal registry & \begin{tabular}[c]{@{}p{8cm}@{}} Exhibiting a well-mannered language registry.\end{tabular}\\

& 20 & Placeholder words & \begin{tabular}[c]{@{}p{8cm}@{}} Use of auxiliary words instead of a more precise one.\end{tabular}\\

& 21 & Sesquipedalian words & \begin{tabular}[c]{@{}p{8cm}@{}} Employing ceremonial, long, uncommon words.\end{tabular}\\

& 22 & Short response & \begin{tabular}[c]{@{}p{8cm}@{}} Providing quick answers.\end{tabular}\\

\bottomrule
\end{tabular}
\end{table*}

\begin{equation}\label{eq:feature_engineering}
\begin{split}
\forall n \in \{1 ... \infty\}\\ \\
X[n] = \{ x[0],\ldots,x[n]\}. \\
Y[n] = \{y_0[n], y_1[n],\ldots,y_{n-1}[n]\} \mid \\ y_0[n]\leq y_1[n]\leq\ldots\leq y_{n-1}[n], \\
\mbox{where} ; \forall x \in X[n], \; x \in Y[n]. \\ \\
avg^n[n]=\frac{1}{n}\sum_{i=0}^{n} y_i [n]\\
Q^n_{1}[n]=y_{\nint{\frac{1}{4}n}}[n] \\
Q^n_{2}[n]=y_{\nint{\frac{2}{4}n}} [n]\\
Q^n_{3}[n]=y_{\nint{\frac{3}{4}n}} [n]\\
\end{split}
\end{equation}

\subsubsection{Feature analysis \& selection}
\label{sec:feature_analysis_selection}

Feature analysis and selection tasks are necessary to optimize the performance of the \textsc{ml} classifiers. These tasks are even more important in the streaming scenario where samples arrive at a real-time pace. The latter means that the classification problem layout (\textit{e.g.}, the most relevant features) may vary over time.

The proposed system follows two thresholding strategies for feature analysis and selection based on cut-off points regarding correlation and variance values to remove irrelevant features. The former, correlation analysis, limits the number of features to extract the most relevant characteristics. For the latter variance analysis, the number of features selected is dynamically established in each interaction of the stream-based model, selecting those that meet the threshold criteria.

Algorithm \ref{alg:data_processing} details the data processing stage, including feature engineering, analysis, and selection.

\begin{algorithm*}[!htbp]
 \caption{\label{alg:data_processing}{\bf Data processing}}
 \begin{algorithmic}[0]
 \Function{$data\_processing$}{$list\_sessions, list\_features, selector\_mode,selector\_threshold$}
    \For{$feature$ $in$ $list\_features$}
        \State $list\_features.append($avg$(list\_sessions[feature]))$
        \State $list\_features.append(Q_1(list\_sessions[feature]))$
        \State $list\_features.append(Q_2(list\_sessions[feature]))$
        \State $list\_features.append(Q_3(list\_sessions[feature]))$
    \EndFor
    \State $list\_features\_selected=[]$
    \For{$feature$ $in$ $list\_features$}
        
        \If {$selector\_mode==``variance"$ and $varizance(feature)>selector\_threshold$}
            \State $list\_features\_selected.append(feature)$
        \ElsIf {$selector\_mode==``correlation"$ and $correlation(feature)>selector\_threshold$}
            \State $list\_features\_selected.append(feature)$
        \EndIf
    \EndFor
    \State $return$ $list\_features\_selected$
 \EndFunction
 \end{algorithmic}
\end{algorithm*}

\subsection{Stream-based classification}
\label{sec:classification}

Two classification scenarios are considered:

\begin{description}
 \item \textbf{Scenario 1} analyzes the behavior of the classifiers in a streaming setting. Under this consideration, sequential and continual testing and training over time is assumed.
 
 \item \textbf{Scenario 2} analyzes the models' performance under more realistic conditions. Thus, the testing is continuous (\textit{i.e.}, in streaming) while training is performed desynchronized in blocks of 100 samples.
\end{description}

The following \textsc{ml} models are selected based on their good performance in similar classification problems \citep{Mathkunti2020,Ilias2023,Kumar2023}:

\begin{itemize}
 \item \textbf{Gaussian Naive Bayes} (\textsc{gnb}) \citep{Xu2018} exploits the Gaussian probability distribution in a stream-based \textsc{ml} model. It is used as a reference for performance analysis.
 
 \item \textbf{Approximate Large Margin Algorithm} (\textsc{alma}) \citep{Kang2019} is a fast incremental learning algorithm comparable to Support Vector Machine to approximate the maximal margin between a hyperplane concerning a norm (with a value of $p \geq 2$) for a set of linearly separable data.
 
 \item \textbf{Hoeffding Adaptive Tree Classifier} (\textsc{hatc}) \citep{Weinberg2023} computes single-tree branch performance and is designed for stream-based prediction.
 
 \item \textbf{Adaptive Random Forest Classifier} (\textsc{arfc}) \citep{Zhang2021} constitutes an advanced model of \textsc{hatc} in which branch performance is computed by majority voting in an ensemble tree scenario.
\end{itemize}

Algorithm \ref{alg:classification} describes the stream-based prediction process.

\begin{algorithm*}[!htbp]
 \caption{\label{alg:classification}{\bf Classification}}
 \begin{algorithmic}[0]
 \Function{$classification$}{$list\_features\_selected, scenario, model\_name, count, list\_y$}
    \State $y\_pred=machine\_learning\_model(model\_name,list\_features\_selected).predict()$

    \If {$scenario==1$}
        \State $machine\_learning\_model(model\_name,list\_features\_selected).train(list\_y[last])$
    \ElsIf{$count\%100 == 0$}
        \State $machine\_learning\_model(model\_name,list\_features\_selected).train(list\_y[-100:last])$  
    \EndIf
    \State $return$ $y\_pred$

 \EndFunction

 \end{algorithmic}
\end{algorithm*}

\subsection{Explainability dashboard}
\label{sec:explainability}

Prediction transparency is promoted through explainability data provided to the end users regarding relevant features in the prediction outcome. Thus, those relevant features are included in the natural language description of the decision path. The five features whose mathematical module is highest or with the highest variance and whose values are the most distant from the average are selected. In the case of the counters (features 9-10), this average is obtained from the average of all users in the system.

\section{Evaluation and discussion}
\label{sec:results}

This section discusses the experimental data set used, the implementation decisions, and the results obtained. The evaluations were conducted on a computer with the following specifications:

\begin{itemize}
 \item \textbf{Operating System}: Ubuntu 18.04.2 LTS 64 bits
 \item \textbf{Processor}: Intel\@Core i9-10900K \SI{2.80}{\giga\hertz}
 \item \textbf{RAM}: \SI{96}{\giga\byte} DDR4 
 \item \textbf{Disk}: \SI{480}{\giga\byte} NVME + \SI{500}{\giga\byte} SSD
\end{itemize}

\subsection{Experimental data set}
\label{sec:dataset}

\begin{figure*}
 \centering
 \begin{subfigure}[b]{0.75\textwidth}
 \centering
 \includegraphics[scale=0.7]{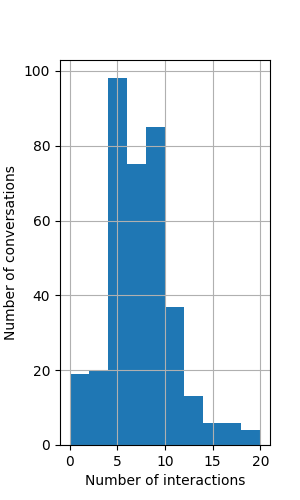}
 \includegraphics[scale=0.7]{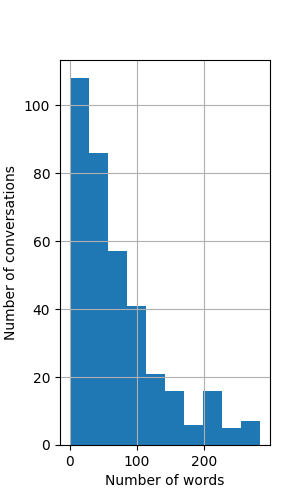}
 \caption{Without mental deterioration.}
 \label{fig:dist_no_cog}
 \end{subfigure}
 \hfill
 \begin{subfigure}[b]{0.75\textwidth}
 \centering
 \includegraphics[scale=0.7]{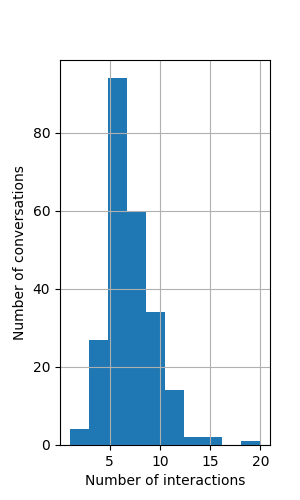}
 \includegraphics[scale=0.7]{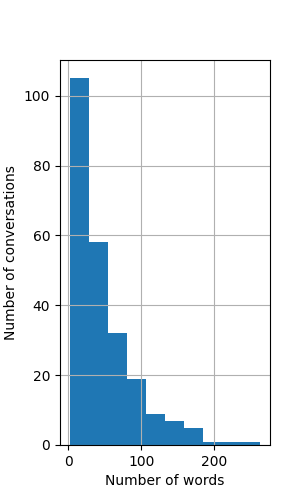}
 \caption{With mental deterioration.}
 \label{fig:dist_cog}
 \end{subfigure}
 \caption{Distribution of interactions and number of words.}
 \label{fig:distribution}
\end{figure*}

The experimental data set\footnote{Data are available on request from the authors.} consists of an average of $6.92\pm3.08$ utterances with $62.73\pm57.20$ words involving \num{44} users with $13.66\pm7.86$ conversations by user. The distribution of mental deterioration in the experimental data set is \num{238} samples in which mental deterioration is present and \num{363} in which it is absent. Figure \ref{fig:distribution} depicts the histogram distribution of words and interactions by absent and present mental deterioration, respectively. While the distributions of the number of interactions in the absence or presence of cognitive impairment follow a normal function, the number of words can be approximated by a positive normal centered on 0. The most relevant issue is that, as expected, users with mental deterioration present a lower number of interactions and a significant decrease in the number of words used in their responses.

\subsection{Data extraction}
\label{sec:feature_extraction_results}

Data to engineer conversational (1-8), emotional, and linguistic features in Table \ref{tab:features} were obtained with \texttt{gpt-3.5-turbo}\footnote{Available at \url{https://platform.openai.com/docs/models/gpt-3-5}, May 2024.} model. The prompt used is shown in Listing \ref{lst:prompt_data_extraction}.

\begin{figure*}[t]
\begin{lstlisting}[frame=single,caption={Prompt used for data extraction.}, label={lst:prompt_data_extraction},emphstyle=\textbf,escapechar=ä]
This is a conversation between a bot and a human. Answer what I ask below with a
value between 0.0 and 1.0, being 0.0 never and 1.0 always.

Detect if the human: has any memory loss, is incoherent, exhibits comprehension 
problems, is confused, fluent, shows initiative, uses repetitive language, hides 
feelings and personal information, expresses mental or physical health concerns, 
is tired, feels lonely, the polarity of the conversation, seems sad, interacts 
with a colloquial registry, has conjugation problems, uses interjections to 
complete pauses, interacts with a formal registry, uses placeholder words, 
sesquipedalian terms, and short responses.

Respond only in the following JSON format: 
{ä``äAmnesia":0.0, ä``Incoherence"ä:0.0, ä``Incomprehension"ä:0.0, ä``Confusion"ä:0.0, ä``Fluency"ä:0.0, 
ä``Initiative"ä:0.0, ä``Repetitiveness"ä:0.0, ä``Secretive"ä:0.0, ä``Health\_state"ä:0.0, ä``Fatigue"ä:0.0, 
ä``Loneliness"ä:0.0, ä``Polarity"ä:0.0, ä``Sadness"ä:0.0, ä``Colloquial\_registry"ä:0.0,
ä``Conjugation\_problems"ä:0.0, ä``Disfluency"ä:0.0, ä``Formal\_registry"ä:0.0, ä``Placeholder\_words"ä:0.0, 
ä``Sesquipedalian words"ä:0.0, ä``Short response"ä:0.0}.

ALWAYS RETURN A JSON IN THE GIVEN FORMAT WITHOUT ADDING MORE TEXT OR MODIFYING 
THE FIELD NAMES IN THE JSON. DO NOT ANSWER ANY QUESTIONS IN THE CONVERSATION.

<Dialogue>

\end{lstlisting}
\end{figure*}

\subsection{Stream-based data processing}
\label{sec:data_processing_results}

This section reports the algorithms used for feature engineering, analysis, and selection and their evaluation results.

\subsubsection{Feature engineering}
\label{sec:feature_engineering_results}

\begin{figure*}[!htpb]
 \centering
 \includegraphics[scale=0.6]{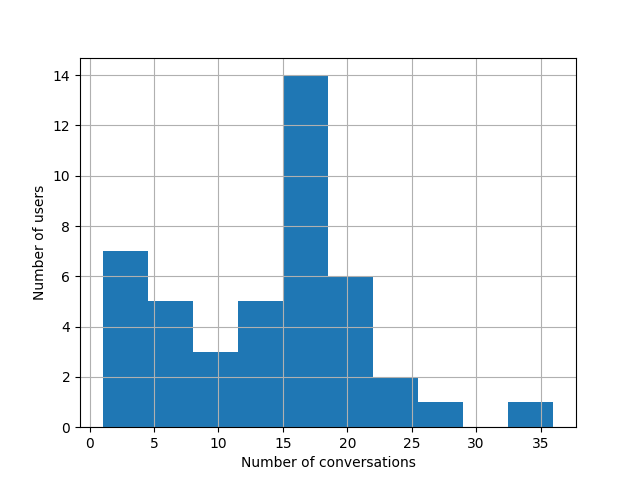}
 \caption{Distribution of conversations by user.}
 \label{fig:conversations_users}
\end{figure*}

A total of 88 features were generated\footnote{New four characteristics (average, \textsc{q}1, \textsc{q}2, and \textsc{q}3) per each of the 22 features in Table \ref{tab:features}.} in addition to the 22 features generated in each conversation (see Table \ref{tab:features}) resulting in 110 features. In Figure \ref{fig:conversations_users}, we show the distribution of conversations by the user, which approaches a uniform density function, being relevant that the large majority concentrates between 15 and 20 conversations.

\subsubsection{Feature analysis \& selection}
\label{sec:feature_analysis_selection_results}

Correlation and variance thresholding decisions were based on experimental tests. For the correlation thresholding, \texttt{SelectKBest}\footnote{Available at \url{https://riverml.xyz/0.11.1/api/feature-selection/SelectKBest}, May 2024.} was applied using the Pearson correlation coefficient \citep{Benesty2009}. The $K$ value corresponds to the most relevant features of the \SI{80}{\percent} experimental data. Table \ref{tab:correlation_variance} shows the features with a correlation value greater than 0.2 with the mental deterioration target when the last sample entered the stream-based classification model.

\begin{table*}[!htpb]
\centering
\caption{\label{tab:correlation_variance}Correlation and variance results.}
\begin{tabular}{cccc}
\toprule
\textbf{Feature} & \begin{tabular}[c]{@{}p{3.5cm}@{}} \centering\textbf{Statistic metric}\end{tabular} & \begin{tabular}[c]{@{}p{3.5cm}@{}} \centering\textbf{Value}\end{tabular} \\
\midrule
\multirow{13}{*}{Correlation} & 22 & Average & 0.296 \\
& \multirow{3}{*}{9} & Q1 & 0.292\\
& & Q2 & 0.248\\
& & Average & 0.219\\
& 19 & Average & -0.203\\
& 18 & Q3 & -0.213\\
& 14 & Q1 & -0.226\\
& 12 & Average & -0.238\\
& \multirow{3}{*}{14} & Q3 & -0.272\\
& & Average & -0.278\\
& & Q2 & -0.318\\
& \multirow{2}{*}{6} & Average & -0.391\\
& & Q3 & -0.458\\
\midrule
\multirow{7}{*}{Variance} & \multirow{2}{*}{16} & Q1 & 0.171 \\
& & Original & 0.165\\
& 6 & Original & 0.11\\
& 16 & Q2 & 0.086\\
& 14 & Original & 0.084\\
& 6 & Q3 & 0.079\\
& 14 & Q3 & 0.055\\
\bottomrule
\end{tabular}
\end{table*}

Regarding the variance thresholding, the implementation used was \texttt{VarianceThreshold}\footnote{Available at \url{https://riverml.xyz/0.11.1/api/feature-selection/VarianceThreshold}, May 2024.} from the \texttt{River} library\footnote{Available at \url{https://riverml.xyz/0.11.1}, May 2024.}. Moreover, the cut-off point, \num{0.001}, is computed with the 10th percentile variance value of the features contained in the \SI{20}{\percent} of the experimental data set, which acts as the cold start of this method. Consequently, only those features that exceed the abovementioned cut-off are selected as relevant for classification purposes. Table \ref{tab:correlation_variance} also details the features with a variance greater than 0.5\footnote{Note that we have discarded features 9 and 10 from Table \ref{tab:features} from this example since they represent counters and their variance is always greater than 1.}.

Table \ref{tab:correlation_variance} shows that among the conversational features, user initiative (feature 6 in Table \ref{tab:features}) plays an important role. The same applies to the number of interactions within a dialogue (feature 9). Regarding emotional features, consideration should be given to fatigue (feature 12) and polarity (feature 14). Finally, using a colloquial/formal registry (features 16/19), disfluency (feature 18), and short responses (feature 22) stand out among linguistic characteristics. Considering correlation and variance analysis jointly, initiative and polarity are the most relevant data for prediction purposes.

\subsection{Stream-based classification}
\label{sec:classification_results}

The River implementations of the \textsc{ml} models selected are: \textsc{gnb}\footnote{Available at \url{https://riverml.xyz/dev/api/naive-bayes/GaussianNB}, May 2024.}, \textsc{alma}\footnote{Available at \url{https://riverml.xyz/0.11.1/api/linear-model/ALMAClassifier}, May 2024.}, \textsc{hatc}\footnote{Available at \url{https://riverml.xyz/0.11.1/api/tree/HoeffdingAdaptiveTreeClassifier}, May 2024.} and \textsc{arfc}\footnote{Available at \url{https://riverml.xyz/0.11.1/api/ensemble/AdaptiveRandomForestClassifier}, May 2024.}. Listings \ref{alma_conf}, \ref{hatc_conf} and \ref{arfc_conf} detail the hyper-parameter optimization ranges used, excluding the baseline model, from which the following values were selected as optimal:

\begin{description}
 \item \textbf{Correlation thresholding}
 \begin{itemize}
 \item \textbf{ALMA}: alpha=0.5, B=1.0, C=1.0.

 \item \textbf{HATC}: depth=None, tiethreshold=0.5, maxsize=50.

 \item \textbf{ARFC}: models=10,features=5, lambda=50.
 \end{itemize}

 \item \textbf{Variance thresholding}
 \begin{itemize}
 \item \textbf{ALMA}: alpha=0.5, B=1.0, C=1.0.

 \item \textbf{HATC}: depth=None, tiethreshold=0.5, maxsize=50.

 \item \textbf{ARFC}: models=100,features=sqrt, lambda=50.
 \end{itemize}
\end{description}

\begin{lstlisting}[frame=single,caption={\textsc{alma} hyper-parameter configuration.},label={alma_conf},emphstyle=\textbf,escapechar=ä]
alpha = [0.5,0.7,0.9]
B = [1.0, 1.41, 1.2]
C = [1.0,1.11, 1.2]
\end{lstlisting}

\begin{lstlisting}[frame=single,caption={\textsc{hatc} hyper-parameter configuration.},label={hatc_conf},emphstyle=\textbf,escapechar=ä]
depth = [None, 50, 200]
tiethreshold = [0.5, 0.05, 0.005]
maxsize = [50, 100, 200]
\end{lstlisting}
 
\begin{lstlisting}[frame=single,caption={\textsc{arfc} hyper-parameter configuration.},label={arfc_conf},emphstyle=\textbf,escapechar=ä]
models = [10, 25, 100]
features = [sqrt, 5, 50]
lambda = [25, 50, 100]
\end{lstlisting}

Table \ref{tab:classification_results} presents the results for evaluation scenarios 1 and 2. In both scenarios, the feature selection methodology based on correlation thresholding returns lower classification metric values than those obtained with the variance method. Thus, once the variance feature selection method is applied, the \textsc{arfc} is the most promising performance algorithm regardless of the evaluation scenario.

\begin{table*}[!htbp]
\centering
\caption{\label{tab:classification_results}Classification results (Sce.: scenario, time in seconds).}
\begin{tabular}{ccccccccccS[table-format=3.2]}
\toprule
\bf Sce. & \bf Selection & \bf Model & \bf Acc. & \multicolumn{3}{c}{\bf Precision} & \multicolumn{3}{c}{\bf Recall} & {\bf Time}\\
\cmidrule(lr){5-8}
\cmidrule(lr){8-10}
& & & & Macro & Present & Absent & Macro & Present & Absent\\
\midrule
\multirow{8}{*}{1} & \multirow{4}{*}{Correlation} 
& \textsc{gnb} & 63.11 & 68.57 & 52.13 & 85.00 & 67.24 & \bf87.39 & 47.09 & 0.76\\
& & \textsc{alma} & 67.67 & 66.20 & 59.32 & 73.08 & 66.15 & 58.82 & \bf73.48 & 0.63\\
& & \textsc{hatc} & 63.27 & 64.65 & 52.69 & 76.60 & 65.09 & 73.95 & 56.23 & 0.98\\
& & \textsc{arfc} & \bf72.29 & \bf74.23 & \bf60.53 & \bf87.94 & \bf74.79 & 86.97 & 62.60 & 1.57\\

\cmidrule{2-11}

& \multirow{4}{*}{Variance} 
& \textsc{gnb} & 61.94 & 59.48 & 56.10 & 62.86 & 54.68 & 19.33 & 90.03 & 0.31\\
& & \textsc{alma} & 67.33 & 65.88 & 58.82 & 72.93 & 65.88 & 58.82 & 72.93 & 0.20\\
& & \textsc{hatc} & 69.78 & 68.98 & 60.52 & 77.44 & 69.63 & 68.91 & 70.36 & 0.54\\
& & \textsc{arfc} & \bf89.15 & \bf88.47 & \bf83.92 & \bf93.02 & \bf89.28 & \bf89.92 & \bf88.64 & 17.72\\

\midrule

\multirow{8}{*}{2} & \multirow{4}{*}{Correlation} 
& \textsc{gnb} & 58.60 & \bf 66.76 & 48.85 & \bf 84.66 & 63.86 & \bf 89.50 & 38.23 & 0.74\\
& & \textsc{alma} & \bf 63.67 & 61.40 & \bf 55.32 & 67.48 & 60.25 & 43.70 & \bf 76.80 & 0.62\\
& & \textsc{hatc} & 58.10 & 55.80 & 47.06 & 64.55 & 55.64 & 43.70 & 67.59 & 0.94\\
& & \textsc{arfc} & 63.27 & 65.44 & 52.57 & 78.31 & \bf 65.66 & 77.31 & 54.02 & 1.42\\

\cmidrule{2-11}

& \multirow{4}{*}{Variance} 
& \textsc{gnb} & 62.10 & 59.86 & 56.79 & 62.93 & 54.82 & 19.33 & \bf 90.30 & 0.31\\
& & \textsc{alma} & 63.33 & 60.89 & 56.00 & 65.78 & 58.53 & 35.29 & 81.77 & 0.20\\
& & \textsc{hatc} & 65.94 & 65.57 & 55.90 & 75.24 & 66.23 & 67.65 & 64.82 & 0.49\\
& & \textsc{arfc} & \bf 84.81 & \bf 84.04 & \bf 78.38 & \bf 89.71 & \bf 84.89 & \bf 85.29 & 84.49 & 15.50\\

\bottomrule
\end{tabular}
\end{table*}

\begin{figure*}[!htpb]
 \centering
 \includegraphics[scale=0.15]{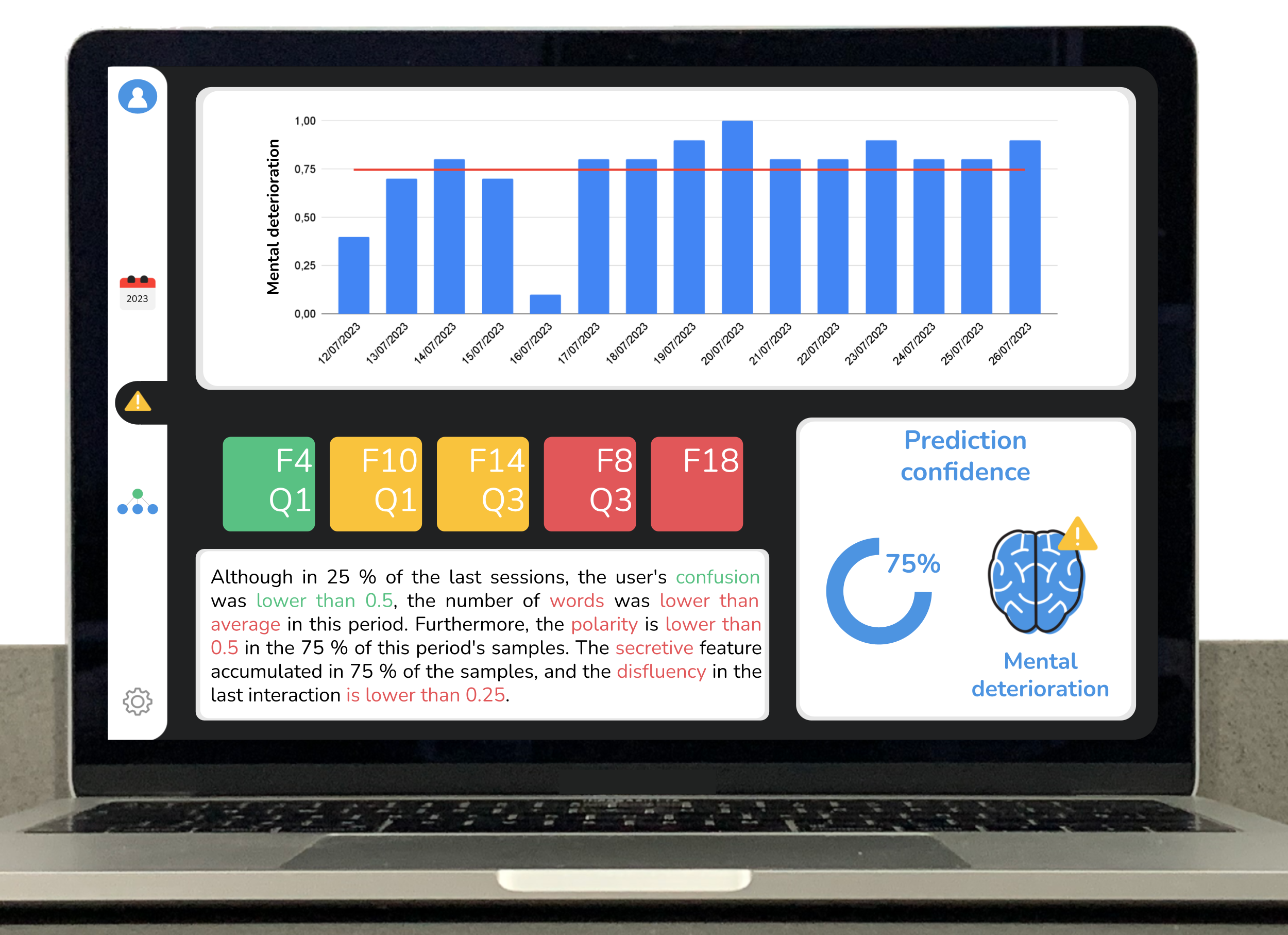}
 \caption{Explainability dashboard.}
 \label{fig:dashboard}
\end{figure*}

Consideration should be given to the fact that even in scenario 2, in which training is performed desynchronized and in batch, the robustness of \textsc{arfc} stands out with classification results exceeding \SI{80}{\percent} and with a recall for the mental deterioration class about \SI{85}{\percent}.

Provided that our system operates in streaming and to enable direct comparison with batch \textsc{ml} solutions, additional evaluation measures from 10-fold cross-validation are provided, particularly, for Random Forest (\textsc{rf}\footnote{Available at \url{https://scikit-learn.org/stable/modules/generated/sklearn.ensemble.RandomForestClassifier.html}, May 2024.}) equivalent to the best model, \textsc{arfc}, in stream-based classification. The results are displayed in Table \ref{tab:classification_results_batch}, most surpassing the \SI{90}{\percent} threshold. Note that the increase in performance compared to streaming operation (\textit{e.g.}, +\SI{8.37}{\percent} points in accuracy) is derived from the fact that in batch classification, the model has access to the \SI{90}{\percent} of the experimental data for training. In contrast, stream-based classification relies on the ordered incoming new samples, which is more demanding. Consequently, having achieved a comparable performance in batch and stream-based classification is noteworthy.

\begin{table*}[!htbp]
\centering
\caption{\label{tab:classification_results_batch}Classification results in batch for the \textsc{rf} model (time in seconds).}
\begin{tabular}{ccccccccS[table-format=3.2]}
\toprule
\bf Acc. & \multicolumn{3}{c}{\bf Precision} & \multicolumn{3}{c}{\bf Recall} & {\bf Time}\\
\cmidrule(lr){2-4}
\cmidrule(lr){5-7}
 & Macro & Present & Absent & Macro & Present & Absent\\
\midrule
93.18 & 93.15 & 93.01 & 93.28 & 92.54 & 89.50 & 95.59 & 1.96\\
\bottomrule
\end{tabular}
\end{table*}

To verify the system's operation in a more challenging scenario, we have experimented with a data set from a previous study \cite{de_Arriba2023} with fewer interactions per session. Even when the system is fed with less information, the evaluation metrics are promising, as shown in Table \ref{tab:classification_results_datasetv2} with all values above \SI{70}{\percent}, and the precision and recall of the mental deterioration category above \SI{80}{\percent}. Comparing the \textsc{rf} batch model in our past research \cite{de_Arriba2023} with the proposed \textsc{arfc} algorithm, which operates in streaming, the improvement reaches \SI{10}{\percent} points and \SI{4}{\percent} points in the recall metric of mental deterioration and absence of mental deterioration categories, respectively.

\begin{table*}[!htbp]
\centering
\caption{\label{tab:classification_results_datasetv2}Classification results for the \textsc{arfc} model using the experimental data from \cite{de_Arriba2023} (time in seconds).}
\begin{tabular}{ccccccccS[table-format=3.2]}
\toprule
\bf Acc. & \multicolumn{3}{c}{\bf Precision} & \multicolumn{3}{c}{\bf Recall} & {\bf Time}\\
\cmidrule(lr){2-4}
\cmidrule(lr){5-7}
 & Macro & Present & Absent & Macro & Present & Absent\\
\midrule
77.70 & 76.62 & 81.32 & 71.93 & 76.46 & 82.22 & 70.69 & 4.72\\
\bottomrule
\end{tabular}
\end{table*}

 \subsection{Explainability dashboard}
\label{sec:explainability_result}

Figure \ref{fig:dashboard} shows the explainability dashboard. In this example, the variation in predicting cognitive impairment is visualized, considering two weeks of past data. This variation is represented with the \texttt{predict\_proba} function of \textsc{arfc} algorithm. At the bottom, the most relevant features are displayed. Each figure card contains the identifier and statistic represented in colors following this scheme: 1 to 0.5 in green, 0.5 to 0.25 in yellow, and 0.25 to 0 in red. The latter assignation is inverted for negative values. At the bottom, a brief description in natural language is provided. The average accumulated \texttt{predict\_proba} value, and the confidence prediction of the current sample are displayed on the right.

\section{Conclusions}
\label{sec:conclusions}

Cognitively impaired users find it difficult to perform daily tasks with the consequent detrimental impact on their life quality. Thus, progression detection and early intervention are essential to effectively and timely address mental deterioration to delay its progress. In this work, we focused on impairment in language production (\textit{i.e.}, lexical, semantic, and pragmatic aspects) to engineer linguistic-conceptual features towards spontaneous speech analysis (\textit{e.g.}, semantic comprehension problems, memory loss episodes, etc.). Compared to traditional diagnostic approaches, the proposed solution has semantic knowledge management and explicability capabilities thanks to integrating an \textsc{llm} in a conversational assistant.

Consideration should be given to the limitations of using \textsc{llm}s, which are transversal into the healthcare field beyond mental deterioration detection. The potential biases and lack of inherent transparency stand out among the risks of applying these models for medical purposes. The latter black-box problem, also present in traditional opaque \textsc{ml} models, is particularly critical in the healthcare field by negatively impacting the decision process of physicians due to their limited corrective capabilities and even the end users, limiting their trust in medical applications. Moreover, these systems' current limited memory management capability is worth mentioning, which prevents the realization of longitudinal clinical analysis. The same applies to the associated complexity of context information management. Ultimately, the difficulty in collecting data due to the sensitivity and confidentiality of the information in the medical field should also be mentioned.

More in detail, the solution provides interpretable \textsc{ml} prediction of cognitive decline in real-time. \textsc{rlhf} (\textit{i.e.}, prompt engineering) and explainability are exploited to avoid the ``hallucination" effect of \textsc{llm}s and avoid potential biases by providing natural language and visual descriptions of the diagnosis decisions. Note that our system implements \textsc{ml} models in streaming to provide real-time functioning, hence avoiding the re-training cost of batch systems. 

Summing up, we contribute with an affordable, flexible, non-invasive, personalized diagnostic system that enables the monitoring of high-risk populations and offers companionship. Ultimately, our solution democratizes access to researchers and end users within the public health field to the latest advances in \textsc{nlp}.

Among the challenges and potential ethical concerns raised by the application of \textsc{ai} into the healthcare field, the \textit{double effect principle} must be considered. In this sense, few can deny its promising potential to provide innovative treatments while at the same time presenting safety-critical concerns, notably regarding their interpretability. Apart from the algorithmic transparency mentioned, the main considerations are privacy and safety of the medical data, fairness, and autonomous decision-making without human intervention. In future work, we plan to test the performance of new approaches, such as reinforcement learning, to enhance the system's personalizing capabilities further. Moreover, we will explore co-design practices with end users, and we seek to move our solution to clinical practice within an ongoing project with daycare facilities. Note that reinforcement learning with human feedback will also allow us to mitigate some of the limitations discussed, such as physicians' lack of interpretability and corrective capabilities. The latter will also have a positive ethical impact on the deployment of \textsc{llm}-based medical applications by ensuring fairness. The societal impact derived from reduced costs compared to traditional approaches may result in broader accessibility to clinical diagnosis and treatment on a demand basis. The equity will be impulsed by the capability of these systems to provide unlimited personalized support. In future research, we will work on mitigating health inequities by performing longitudinal studies to measure bias in our \textsc{ai} solution, particularly related to the algorithm design, bias in the training data, and the ground truth. Underperformance in certain social groups may also be considered. For that purpose, we will gather social context data, which will allow us to measure equity (\textit{e.g.}, gender, race, socioeconomic status, etc.). To ensure patient data protection while at the same time increasing data available for research, federated learning approaches will be explored.

\section*{Declarations}

\subsection*{Competing interests}

The authors have no competing interests to declare relevant to this article's content.

\subsection*{Funding}

This work was partially supported by (\textit{i}) Xunta de Galicia grants ED481B-2022-093 and ED481D 2024/014, Spain; and (\textit{ii}) University of Vigo/CISUG for open access charge.

\subsection*{Authors contribution}

\textbf{Francisco de Arriba-Pérez}: Conceptualization, Methodology, Software, Validation, Formal analysis, Investigation, Resources, Data Curation, Writing - Original Draft, Writing - Review \& Editing, Visualization, Supervision, Project administration, Funding acquisition. \textbf{Silvia García-Méndez}: Conceptualization, Methodology, Software, Validation, Formal analysis, Investigation, Resources, Data Curation, Writing - Original Draft, Writing - Review \& Editing, Visualization, Supervision, Project administration, Funding acquisition.

\bibliography{bibliography}

\end{document}